# Joint Distribution Alignment via Adversarial Learning for Domain Adaptive Object Detection

Bo Zhang, Tao Chen, Bin Wang, *Senior Member, IEEE,* Ruoyao Li

*Abstract*—Unsupervised domain adaptive object detection aims to adapt a well-trained detector from its original source domain with rich labeled data to a new target domain with unlabeled data. Recently, mainstream approaches perform this task through adversarial learning, yet still suffer from two limitations. First, they mainly align marginal distribution by unsupervised cross-domain feature matching, and ignore each feature's categorical and positional information that can be exploited for conditional alignment; Second, they treat all classes as equally important for transferring cross-domain knowledge and ignore that different classes usually have different transferability. In this paper, we propose a joint adaptive detection framework (JADF) to address the above challenges. First, an end-to-end joint adversarial adaptation framework for object detection is proposed, which aligns both marginal and conditional distributions between domains without introducing any extra hyper-parameter. Next, to consider the transferability of each object class, a metric for class-wise transferability assessment is proposed, which is incorporated into the JADF objective for domain adaptation. Further, an extended study from unsupervised domain adaptation (UDA) to unsupervised few-shot domain adaptation (UFDA) is conducted, where only a few unlabeled training images are available in unlabeled target domain. Extensive experiments validate that JADF is effective in both the UDA and UFDA settings, achieving significant performance gains over existing state-of-the-art cross-domain detection methods.

*Index Terms*—Joint adaptive detection framework, domain adaptation for object detection, adversarial learning, class-wise transferability.

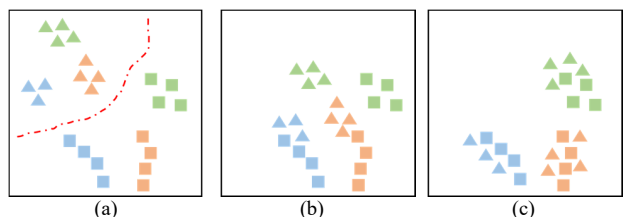

Fig.1. Illustration of different domain adaptation methods for a classification task. Different colors denote features from different classes. The square and triangle represent features from the source domain and target domain, respectively. (a) Features from source and target domains are separated by a red dashed line denoting the boundary between the two domains, without domain adaptation. (b) Features from the two domains are marginally adapted, by unsupervised cross-domain feature matching ignoring each feature's categorical attribute. (c) Features from the two domains are jointly adapted by complete adaptation, which considers both marginal adaptation and conditional adaptation that exploits each feature's categorical information

## I. INTRODUCTION

ADAPTING a well-trained object detector from its source domain to a new target domain is important, especially when the annotation of images on the new domain is very scarce or the cost of image acquisition is high [1]-[3]. However, due to data distribution difference between source and target domains, such a domain adaptation task is challenging, and has motivated a lot of research works [4]-[8], which enable data-rich source knowledge to be transferred to a new domain.

For cross-domain object detection, the source and target images can be considered to be generated by sampling from different probability distributions. Therefore, domain adaptive detector learning mainly focuses on how to mitigate the impact of data distribution difference between domains on the detector performance [9]. With the rapid development of convolutional neural networks (CNNs) in computer vision field, recent studies employ various CNN variants to solve this task, which can be categorized into two classes: 1) pixel-level adaptation [10]-[13] and 2) feature-level adaptation [14]-[21]. The former generates images similar to the target domain and retrains the detector using the generated images. However, it is difficult to ensure that the generated images have similar pixel-level distribution to that in the target domain [22]. Besides, both the image-to-image translation and detector retraining impose heavy computational burden on the rapid adaptation of well-trained source detectors on the target domain [23]. For the feature-level adaptation, it aims to develop various CNN models to reduce the feature gap between different domains and is easier to be embedded into an adaptive detection framework. However, these works still suffer from two limitations as follows.

Firstly, these works [14]-[21] simply align source and target features without considering each feature's categorical label. As a result, the aligned features become insensitive to different classes and have poor discriminative capability on the target domain. In other words, they do not simultaneously align the marginal and conditional probability distributions, which correspond to cross-domain feature matching in unsupervised and supervised ways, respectively. This is well demonstrated in Fig. 1, which shows the domain adaptation process for a typical classification task. It can be seen from Fig. 1(a) and Fig. 1(b) that when the gap between two domains is large, only aligning the marginal distribution cannot guarantee the optimal adaptation. This case will become even worse for object

This manuscript was first submitted on Dec. 30, 2020 for review. This work was supported by the National Natural Science Foundation of China under Grant No. 61731021. *(Corresponding author: Bin Wang)*

The authors are with the Key Laboratory for Information Science of Electromagnetic Waves (MoE), Fudan University, Shanghai 200433, China, and also with the Research Center of Smart Networks and Systems, School of Information Science and Technology, Fudan University, Shanghai 200433, China (e-mail: wangbin@fudan.edu.cn).

detection task, which needs to consider both object classification and localization for feature alignment. Fig. 1(c) shows that after aligning both marginal and conditional distributions, better adaptation with closer cross-domain feature matching can be achieved.

Secondly, they [14]-[21] treat all object classes as equally important for transferring cross-domain knowledge and ignore that different classes usually have different transferability. The reason is that different objects often present different domain semantic gaps. For some classes with smaller domain gaps, they are relatively easier to be adapted to target domain. For some classes with larger domain shifts, their cross-domain alignment should be weakened.

Accordingly, it is important to consider both the marginal and conditional probability distributions as well as class-wise transferability for joint domain adaptation. In this work, we propose a novel joint adaptive detection framework (JADF). The JADF, which can be applied to a variety of object detection networks including SSD, RefineDet, etc., is mainly composed of a selected baseline detector network, a designed marginal adaptation module and a developed conditional adaptation module considering class-wise transferability of instance objects. Especially, a domain classifier is created to quantify the marginal (or conditional) distribution difference between two domains, and further integrated with adversarial learning to reduce the feature representation gap for the same object. Besides, to align different classes of instance objects according to their different inter-domain shifts, a class-wise $\mathcal{H}$-divergence metric is designed to quantify the transferability of each class. Finally, we extend the study from unsupervised domain adaptation (UDA) to unsupervised few-shot domain adaptation (UFDA), which refers to doing domain adaptation under 1) unsupervised setting which means that all the images from the target domain are unlabeled and 2) few-shot setting which means that only a few training images are available in the unlabeled target domain.

The main contributions of this paper can be concisely summarized as follows:

1) We propose a jointly unsupervised domain adaptation detection framework with end-to-end adversarial learning, which does not introduce any additional hyper-parameter. This is a novel work to consider both marginal and conditional distribution alignments for cross-domain object detection.

2) We study the class-wise transferability problem in the domain adaptation field, and propose a transferability assessment metric and further incorporate it into the JADF optimization objective to consider the transferability of each class during the adaptation process. To our knowledge, this is the first work to investigate class transferability difference for UDA.

3) We extend the study from UDA to UFDA where only a few images from each class on the target domain are given. Extensive validation from UDA to UFDA is conducted, and it is shown that the proposed JADF works well not only when a large number of unlabeled training samples from the target domain are available, but also when only a few unlabeled samples from the target domain are given.

The remainder of this paper is organized as follows: Section II briefly reviews the works on object detection, domain adaptation, and domain adaptation for object detection. In Section III, a probabilistic perspective of joint distribution adaptation is firstly presented, then the joint adaptive detection framework is proposed and its network structure is shown in details, and finally, the optimization objectives and joint adaptation strategy of the proposed method are given. In Section IV, the proposed method on different cross-domain scenarios is evaluated, and the insightful analyses are given. Concluding remarks are presented in Section V.

## II. RELATED WORKS

### A. Object Detection

Benefiting from the rapid development of CNNs in computer vision study, the object detection research has been pushed forward a lot [24]-[26]. These CNN-based object detectors can be roughly categorized into two classes: region-based (two-stage) detectors [27]-[32] and region-free (single-stage) detectors [33]-[39]. The typical region-based detectors, such as RCNN [27], Fast-RCNN [28], Faster-RCNN [29], FPN [30], R-FCN [21], divide the detection process into two stages: proposal generation and instance refinement. One key component is the region-of-interest (RoI) pooling operation [32], which maps each generated proposal with different shapes to the backbone feature maps and pools them into fixed-size features for subsequent refinement. On the other hand, the region-free family, including SSD [33], YOLO [34], YOLO-V2 [35], YOLO-V3 [36], RefineDet [37], detects instance objects directly from the image in a single stage, saving the ROI-pooling operations and leading to significant speed-up in gradient backpropagation and forward inference [38], [39].

Region-based detectors usually have higher detection accuracies than region-free detectors due to the two-stage proposal generation and refinement process, but their inference speed is lower compared to single-stage inference. Moreover, both region-based and region-free detectors still face substantial performance degradation when applied to new data domains, since these detectors only have poor capability to adapt from one domain to new domains with different data distributions.

### B. Domain Adaptation

Unsupervised domain adaptation has been widely investigated in the image classification task [1]-[9], [40]-[45]. Common methods of domain adaptation align feature distributions by reducing the distribution discrepancy in the feature space using two main technologies, including moment matching [1], [40]-[41] and adversarial training [42]-[45].

The moment matching methods aim to align the first-order moments [1], [40] and the second-order moments [41] of two different feature distributions, by means of maximum mean discrepancy (MMD) and deep correlation alignment. On the other hand, more recent works [42]-[45] start to explore the unsupervised distribution matching method by leveraging the concept of adversarial learning, where the adversarial adaptation network consists of two parts: a feature extractor that can learn discriminative features from labeled source data, and a domain classifier that can predict the domain labels. However, most researches in domain adaptation field only focus on the image classification task, while far less progress has been made in the field of object detection, since more challenging problems must be considered for simultaneous optimizations of both object classification and localization for feature alignment.

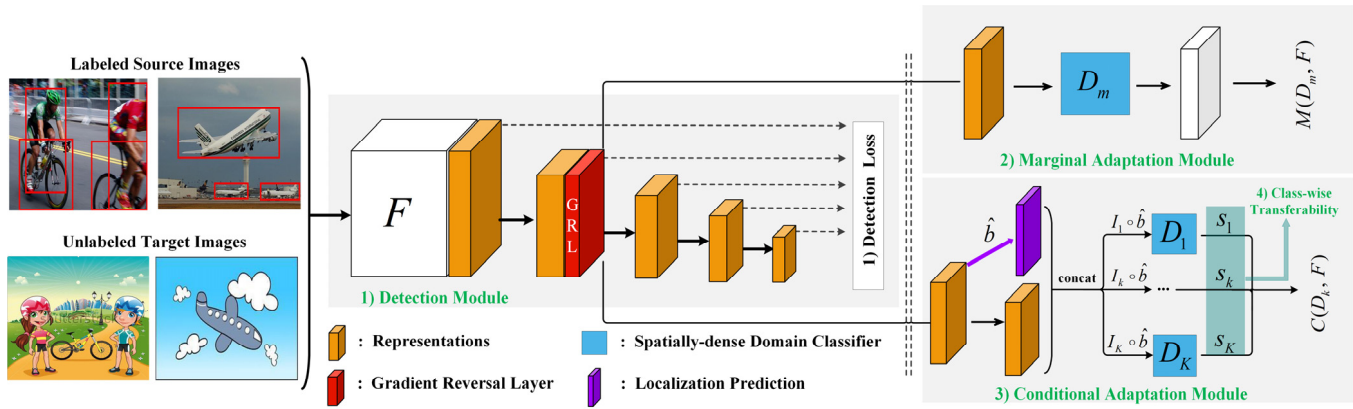

Fig. 2. The overview of JADF. SSD [33] is selected as the baseline detector. We employ the gradient reversal layer (GRL) in [45] to implement adversarial training process. Our implementation aligns the fc7 layer of SSD. Red bounding boxes in the source images represent the ground truth. $I_k$ and $S_k$ denote the feature representations and the transferability of the class $k$, respectively, and $\hat{b}$ is the prediction values of position label.

### C. Domain Adaptation for Object Detection

To investigate how to equip an object detector with good domain adaptation capability, we have reviewed existing domain adaptive object detection works [10]-[21] and find that they are mainly based on unsupervised methods. According to their adaptation ways, these works can be divided into two classes: pixel-level adaptation [10]-[13] and feature-level adaptation [14]-[21].

For the pixel-level adaptation methods, they try to generate new images following the data distribution of the target domain by an image-to-image translation model, and then retrain the detector model using the new labeled images. Such an adaptation method can alleviate the marginal and conditional distribution discrepancy problem, on condition that the generated new images have good quality and high-fidelity, and strictly follow the data distribution on the target domain. However, this is difficult to be guaranteed, since the training of GAN [22], [23] often suffers from mode collapse, especially when only a few target images are given.

For the feature-level adaptation methods, they focus on learning a domain adaptive detector by aligning cross-domain feature representations. An initial attempt [14] is to embed the adversarial domain classifier into the Faster-RCNN [29] to achieve both image-level and instance-level alignment. Such an attempt is extended by the hierarchical alignment module [15], self-training method [16] and strong and weak alignment of features [17] to further improve the adaptability of detectors. Recent works propose to match crucial image regions or important instances across domains by means of categorical regularization [20] and discriminative region mining [21] methods. However, when aligning the features, these domain adaptive detection methods [14]-[21] ignore the categorical attribute of each feature and try to perform cross-domain feature matching in an unsupervised manner. Such marginal adaptation may cause mismatch between different classes of features across domains. Considering that object detection requires both classification and localization, the categorical and positional attributes residing in a feature should be exploited for feature alignment to achieve joint marginal and conditional adaptation. Our work focuses on this joint adaptation, and tries to develop a novel approach with the aid of adversarial learning for UDA and UFDA.

### III. THE PROPOSED METHOD

The purpose of this work is to transfer a pre-trained detector on the source domain to an unlabeled target domain under both UDA and UFDA settings. The overall framework of JADF is shown in Fig. 2. JADF consists of four parts: a baseline detection module, a marginal adaptation module, a conditional adaptation module, and an approach for class-wise transferability assessment. For easy understanding, we first give the probabilistic formulation of the problem. Then we present the JADF and the detailed network structure. Finally, we give the overall optimization objective and adaptation strategy of JADF.

### A. Probabilistic Perspective

*1) Problem Definition:* Suppose that $x$ is an image, $I$ is its feature representation where $I = F(x)$ and $F$ is a learned feature extraction backbone network, and $y$ and $b$ are the category and position labels of an object in the image, respectively. We consider data from the source domain $s$ and target domain $t$ as sampled from different joint probability distributions $P_s(y_s, b_s, I_s)$ and $P_t(y_t, b_t, I_t)$, respectively, i.e., $P_s(y_s, b_s, I_s) \neq P_t(y_t, b_t, I_t)$. The purpose of domain adaptation is to learn a generalized $F$ between domains such that $P_s(y_s, b_s, I_s) = P_t(y_t, b_t, I_t)$.

*2) Joint Adaptation:* According to Bayes theorem, the joint distribution can be formulated as follows:

$$P(y, b, I) = P(I | y, b)P(y, b) = P(y, b | I)P(I). \quad (1)$$

It can be seen that in order to align $P(y, b, I)$, we have two alternatives: aligning $P(I | y, b)P(y, b)$ or aligning $P(y, b | I)P(I)$. For the former case, since the object annotation distribution $P(y, b)$ is actually inconsistent between domains and unable to be aligned, it is difficult to align the joint distribution provided that only $P(I | y, b)$ is aligned. Therefore, we resort to align both the marginal probability distribution $P(I)$ and conditional probability distribution $P(y, b | I)$ simultaneously.

### B. Joint Adaptive Detection Framework

To better illustrate the joint probability distribution alignment theory, we use the region-free detector SSD [33] as the baseline detector, and develop a novel marginal and conditional adaptation approach to construct an end-to-end joint domain adaptive detection framework.

*1) Detection Module:* Given images $x_s$ from the source domain and the feature extractor (backbone) $F$, the detection objective is to optimize $F$ such that the loss $L_{det}(F)$ is minimized:

$$L_{det}(F) = \mathbb{E}_{(y_s, b_s, I_s) \sim P_s} [L_{cls}(F(x_s), y_s) + L_{loc}(F(x_s), b_s)] \quad (2)$$

where $L_{cls}$ and $L_{loc}$ are the classification and localization losses, respectively.

The purpose of UDA for object detection is to learn a good $F$ so that the loss $L$ can reach the minimum for samples from both the source and target domains, by imposing alignment requirement for $P_s(y_s, b_s, I_s) = P_t(y_t, b_t, I_t)$. This requirement can be further converted to simultaneously ensure $P_s(I_s) = P_t(I_t)$ and $P_s(y_s, b_s | I_s) = P_t(y_t, b_t | I_t)$ according to the above analysis.

*2) Marginal Adaptation Module:* Aligning marginal distribution $P(I)$ between domains is challenging, since source domain data suffer from large variations, e.g., in image style, instance object appearance and image scales, which tend to cause target domain image representations $I_t = F(x_t)$ to marginally deviate from the data distribution of source domain, i.e., $P_s(I_s) \neq P_t(I_t)$.

To solve this, we design the following adversarial domain classifier to measure the above cross-domain distribution deviations, and optimize it to reduce the gap:

$$M(D_m, F) = \mathbb{E}_{I_s \sim P_s} [\log(D_m(F(x_s)^{(u,v)}))] \\ + \mathbb{E}_{I_t \sim P_t} [\log(1 - D_m(F(x_t)^{(u,v)}))] \quad (3)$$

where $F(x)^{(u,v)}$ is the feature located at $(u, v)$ of the feature map, and $D_m$ denotes the domain classifier designed to measure the marginal inconsistency. The function of the domain classifier is to distinguish features from the source or target domains. Besides, considering that object detection requires predicting multiple instance objects from an input image, we employ a spatially-dense domain classifier based on $F(x)^{(u,v)}$ to perform patch-level feature adaptation. Essentially it can realize the expansion of the number of training samples from the target domain so as to alleviate the problem of data scarcity.

Note that the marginal adaptation is a kind of unsupervised cross-domain feature matching, which does not consider the category and position labels. Such alignment does not consider decision boundary information, and may cause *uneven/biased adaptation*, resulting in poor discriminative capability for the target domain. On the other hand, only doing marginal adaptation for object detection will result in ignoring cross-domain transferability differences for different classes of objects, which may cause the so-called *negative transfer*, meaning that some non-transferable objects are over-adapted. Therefore, we further propose conditional adaptation considering class-wise transferability to solve the above problems.

*3) Conditional Adaptation Module:* As analyzed before, not only need to align $P(I)$, we also need to further align the posterior probability $P(y, b | I)$ across domains. Here we resort to Bayes Formula $P(y, b | I) = P(I | y, b)P(y, b) / P(I)$ to align $P(I | y, b)P(y, b)$. However, for the unsupervised scenario, the annotation $(y_t, b_t)$ and its distribution $P_t(y_t, b_t)$ in the target domain are unknown. One feasible solution is to utilize the source domain detector to predict object category and position labels for the target domain. Such prediction would gradually mimic the true distribution of the target domain during the detector's iterative updating process. The conditional probability on the target domain can then be transformed as follows:

$$P_t(y_t, b_t | I_t) = \frac{P_t(I_t | y_t, b_t) P_t(y_t, b_t)}{P_t(I_t)} \\ \approx \frac{P_t(I_t | \hat{y}_t, \hat{b}_t) P_t(\hat{y}_t, \hat{b}_t)}{P_t(I_t)} \quad (4)$$

Moreover, to keep consistent and make the conditional adaptation module easy to be integrated into the detection task, we also predict the object category and position information for the source domain. The conditional probability on the source domain can be transformed as follows:

Joint Distribution Alignment via Adversarial Learning for DAOD 5

$$P_s(y_s, b_s \mid I_s) = \frac{P_s(I_s \mid y_s, b_s) P_s(y_s, b_s)}{P_s(I_s)} \approx \frac{P_s(I_s \mid \hat{y}_s, \hat{b}_s) P_s(\hat{y}_s, \hat{b}_s)}{P_s(I_s)} \quad (5)$$

Note that the object category and position labels $(\hat{y}_t, \hat{b}_t)$ for the target domain are predicted by the source detector, and they should have the same probability distribution as $(\hat{y}_s, \hat{b}_s)$ for the source domain. Thus, we take $P_t(\hat{y}_t, \hat{b}_t) = P_s(\hat{y}_s, \hat{b}_s)$.

Here, $P_t(I_t)$ equals $P_s(I_s)$ as guaranteed by marginal adaptation, and $P_t(\hat{y}_t, \hat{b}_t)$ equals $P_s(\hat{y}_s, \hat{b}_s)$ as the annotations are predicted by the same source detector. Hence our goal is converted to align $P(I \mid y, b)$ to promote the matching of conditional distribution. To do this, JADF employs multiple spatially-dense conditional domain classifiers to compute the class-wise domain divergence as follows:

$$L_k(D_k, F) = \mathbb{E}_{I_s \sim P_s}[\log D_k(\hat{y}_{k,s}^{(u,v)}(F(x_s)^{(u,v)} \circ \hat{b}_{k,s}^{(u,v)}))] + \mathbb{E}_{I_t \sim P_t}[\log(1 - D_k(\hat{y}_{k,t}^{(u,v)}(F(x_t)^{(u,v)} \circ \hat{b}_{k,t}^{(u,v)})))] \quad (6)$$

where $L_k$ and $D_k$ denote the adversarial loss and conditional domain classifier w.r.t. the class $k$ ($k \in \{1,...,K\}$) in the category label set $Y$, respectively. Besides, $\hat{y}_k^{(u,v)}$ and $\hat{b}_k^{(u,v)}$ are the category and position offset predictions of the class $k$ located at $(u,v)$, respectively, and "$\circ$" denotes the concatenation operation.

*4) Class-wise Transferability Assessment:* As analyzed before, different classes of instance objects have different cross-domain gaps, and such fact should be considered when designing the conditional domain adaptation module. Inspired by the theory [46] which utilizes the $\mathcal{H}$-divergence to measure the divergence between two sets of samples with different distributions, we design a class-wise divergence metric $d_{k,\mathcal{H}}(I_s, I_t)$ to measure each object class's transferability between domains as follows:

$$d_{k,\mathcal{H}}(I_s, I_t) = 2[1 - \min_{D_k \in \mathcal{H}}(\varepsilon_{k,s}(D_k(I_{k,s})) + \varepsilon_{k,t}(D_k(I_{k,t})))] \quad (7)$$

where $\varepsilon_{k,s}$ and $\varepsilon_{k,t}$ denote the generalization error of the conditional domain classifier $D_k$ w.r.t. the class $k$ on the source and target representations, respectively. $\mathcal{H}$ is the set of possible functions.

It can also be observed from the above metric that the generalization error (prediction loss) of the conditional domain classifier is inversely proportional to each class's cross-domain distance. Thus, we use a class-specific prediction loss $s_k$ from each conditional domain classifier $D_k$ to represent transferability of the class $k$ across domains, and weight $L_k(D_k, F)$ to get the final conditional adaptation objective as follows:

$$C(D_k, F) = \sum_{k=1}^{K} s_k L_k(D_k, F). \quad (8)$$

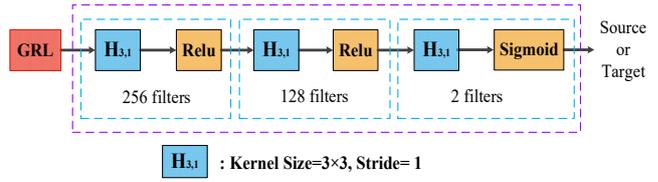

Fig. 3. The network structure of domain classifiers $D_m$ and $D_k$.

### C. Detailed Network Structure of Domain Classifier

Both $D_m$ and $D_k$ employ a three layer Multi-Layer Perceptron (MLP) network with Relu activation function for the first two layers. Besides, we insert a Sigmoid activation function at the end of $D_m$ and $D_k$ since the purpose of the domain classifier is to distinguish features of instance objects from the source domain or target domain. The visual network structure of our designed domain classifier is illustrated in Fig 3.

### D. Overall Objectives and Adaptation Strategy

The two adaptation modules in JADF inherently promote each other. Firstly, the marginal adaptation can properly align feature distributions between domains, which is beneficial for the conditional adaptation to obtain reliable predictions of category label and position offset for the target domain. Secondly, the optimization of conditional adaptation can alleviate the *uneven/biased adaptation* problem, which is caused by the ignorance of decision-making boundaries between object classes during marginal adaptation, and it is also beneficial to improve marginal adaptation. Thirdly, since the object annotations for the target domain are unavailable, we have to utilize the source detector to predict them, which may result in large deviations when the domain shifts are high. The class-wise transferability assessment is designed for assigning small weights $s_k$ in Eq. (8) to some hard-to-transfer object classes with large domain shifts. Consequently, the cross-domain alignment of these hard-to-transfer classes will be weakened by the class-wise transferability assessment.

*1) Optimization Objectives:* The overall loss function of JADF can be written as follows:

$$\min_F L_{det}(F), \quad (9)$$

$$\max_{\{D_m, D_k\}} \min_F L_{det}(F) - \lambda(M(D_m, F) + C(D_k, F)) \quad (10)$$

where $\lambda$ is set to 1.0, and our method does not introduce any additional hyper-parameter.

*2) Adaptation Strategy:* To transfer the learned knowledge from label-rich source domain to an unlabeled target domain, we utilize the following two-stage adaptation strategy to train the JADF model.

**Pre-training phase:** Only the baseline detection model is trained using data from the source domain by Eq. (9). This ensures that the detector can learn sufficient knowledge for subsequent transfer.

**Joint adaptation phase:** The labeled source images and unlabeled target images are mixed together for jointly cross-domain alignment by Eqs. (9) and (10). The marginal and conditional adaptations are optimized in an iterative manner. When the change of the adversarial loss tends to be stable and has less fluctuation, the optimization process ends.

## IV. EXPERIMENTAL RESULTS AND ANALYSES

In this section, we first describe the datasets used to evaluate the proposed JADF. Then, we give the experimental setup for both the pre-training and the joint adaptation phases. Further, we evaluate JADF under two domain adaptation settings: 1) UDA which means that all unlabeled images from the train set of the target domain are available; 2) UFDA which means that only a few unlabeled images (1-shot, or 2-shot, or 3-shot) from the train set of the target domain are available. Finally, we give insightful analyses of the proposed JADF.

### A. Datasets

Six public datasets are used to evaluate the cross-domain detection capability of the proposed JADF, including PASCAL VOC [47], Clipart [12], Comic [12], Watercolor [12], Cityscapes [48], and FoggyCityscapes [49], respectively.

*1) PASCAL VOC:* This dataset mainly covers a large number of real-world images containing 20 categories. For some typical cross-domain detection scenarios [12], [47], PASCAL VOC is usually used as the source domain.

*2) Clipart, Comic, and Watercolor:* These datasets are recently proposed to investigate the domain adaptability of detectors under different image styles, which cover graphical images, cartoon images, and painting, respectively. Artistic or hand-painted images in these datasets have large domain gaps with real-world images in PASCAL VOC. For the typical cross-domain detection scenarios [12], [47], these datasets are generally used as the target domain.

*3) Cityscapes:* This dataset collects a larger number of high-quality images from outdoor street scenes under normal weather conditions, containing 8 categories that often appear around the outdoor street.

*4) FoggyCityscapes:* Unlike the Cityscapes dataset that captures some images under the normal weather condition, FoggyCityscapes simulates the foggy weather condition with different visibility ranges.

### B. Design of Baseline Models

To make a comprehensive comparison with current cross-domain detection models, we first employ SSD [33] as our baseline detector. Next, we devote to study the impact of aligning the joint distribution on common object detection networks, and design several models including SSD+M (denoting SSD with the marginal alignment), SSD+C (denoting SSD with the developed conditional alignment), SSD+M+C (denoting SSD with the marginal alignment and the developed conditional alignment), and SSD+M+WC (denoting SSD with both the marginal alignment and the conditional alignment considering the class-wise transferability). Among the above models, JADF *refers to* the SSD+M+WC, which assembles all developed modules. All the SSD-based models use $300 \times 300$ input size.

### C. Experimental Setup

We employ the two-stage adaptation strategy to train all models, and evaluate our method using mean Average Precision (mAP) under the threshold of 0.5:

*1) Pre-training Phase:* To learn sufficient transferable knowledge for the subsequent domain adaptation process, we first perform the pre-training phase. We use the pre-trained model of VGG-16 on ImageNet [51] and default experimental setups in [33]. Specifically, an initial learning rate of 0.001 is employed in the first 80000 iterations, and the learning rate decays to 0.0001 and 0.00001 at 100000 and 120000 iterations, respectively. The source domain model is trained using SGD with a minibatch size of 32, momentum of 0.9, and weight decay of 0.0005.

During this phase, only labeled images from the source domain are given. For adaptation from PASCAL VOC to Clipart, Comic, Watercolor, we follow the common official split lists in [16], [33] and [37], which adopt the *trainval* list of both VOC07 (containing 5011 images) and VOC12 (containing 11540 images) as the training images, and employ the *test* list of VOC07 (containing 4952 images) as the test set.

*2) Joint Adaptation Phase:* For the joint adaptation process, we insert the marginal adaptation module, the conditional adaptation module, and the class-wise transferability assessment into the detection network. Then we fine-tune the pre-training model on the mixture set that is composed of labeled source images and unlabeled target images. The learning rate is set to 0.0001, and then divided by 10 after the first 300 iterations. The total adaptation process is finished when 600 iterations are reached. After domain adaptation is achieved, all the domain classifiers are removed during the inference phase. Hence our method brings no computational cost to the baseline detector for testing use.

**UDA Setting:** In order to accord with the setups in [16], [17], and [20], 1000 images from Clipart are used during both the joint adaptation phase (only unlabeled images are given) and the evaluation phase. For adaptation from PASCAL VOC to Comic and Watercolor, we use the official *train* list (1000 unlabeled images) as the train set on the target domain, and report their results on the official *test* list (1000 images).

**UFDA Setting:** In the unsupervised few-shot setting, we randomly select $n$ unlabeled images per class from the unlabeled target domain to train the JADF model. In the ablation study, we evaluate JADF by varying $n$ from 1 to 3, repeat each experiment 10 times, and report their means and variances.

TABLE I
ADAPTATION RESULTS FROM PASCAL VOC TO CLIPART. " \ " MEANS THAT THE RESULTS ARE NOT GIVEN IN THEIR ORIGINAL PAPER. " * " MEANS THAT WE RE-IMPLEMENTED THE METHOD. M AND C DENOTE THE MARGINAL AND THE CONDITIONAL ADAPTATION MODULES, RESPECTIVELY. WC DENOTES THE CONDITIONAL ADAPTATION CONSIDERING THE CLASS-WISE TRANSFERABILITY. NOTE THAT WE RE-IMPLEMENTED TWO VERSIONS OF SW-FASTER [17] IN UFDA SETTING, WHICH EMPLOYS VGG-16 AND RESNET [52] AS THE BACKBONE NETWORK, RESPECTIVELY.

| | Method | UDA Setting | | UFDA Setting | | FPS |
|---|---|---|---|---|---|---|
| | | mAP on Target | mAP on Source | mAP on Target | mAP on Source | |
| Single-stage Detector | SSD [33] | 27.6 | **77.5** | 27.6 | **77.5** | **46.0** |
| | SSD+M (ours) | 34.3 | 77.6 | 33.2±0.8 | 77.5±0.2 | |
| | SSD+C (ours) | 35.5 | 76.4 | 34.6±0.7 | 76.8±0.4 | |
| | SSD+WC (ours) | 36.7 | 76.8 | 36.2±0.7 | 76.4±0.3 | |
| | SSD+M+C (ours) | 38.5 | 76.8 | 37.4±0.3 | 76.8±0.4 | |
| | SSD+M+WC (ours) | **39.9** | 76.8 | **38.5±0.2** | 76.6±0.1 | |
| | DT-SSD [12] | 38.0 | \ | \ | \ | |
| | WST-BSR-SSD [16] | 35.7 | \ | \ | \ | |
| | ADDA-SSD [50] | 27.4 | \ | \ | \ | |
| Two-stage Detector | Faster [29] | 27.8 | 77.5 | 27.8 | 77.5 | 2.4 |
| | DA-Faster [14] | 28.5* | 74.5* | 27.3±1.1* | 74.5±1.2* | 2.4 |
| | SW-Faster-VGG-16 [17] | 31.5 | 70.3 | 30.9±0.8* | 70.9±0.4* | 7.0 |
| | SW-Faster-ResNet [17] | 38.1 | 77.0 | **35.9±1.4*** | 74.7±0.3* | 2.4 |
| | SW-Faster-ICR-CCR [20] | **38.3** | \ | \ | \ | 2.4 |

*D. Experimental Results*

Since several recent SSD-based works [12], [16], [50] have conducted the experiments for the cross-domain detection scenarios from PASCAL VOC to Clipart, Comic, and Watercolor, respectively, we follow these SSD-based works and also use these datasets to evaluate our JADF under the UDA and UFDA settings. Next, we consider a more practical detection scenario: given some street images widely collected under the normal weather condition, the domain adaptability of the JADF model under different weather conditions is also reported.

*1) PASCAL VOC to Clipart:* As shown in Table I, our method can significantly improve the mAP of all baseline detectors without using any annotations on the target domain. The results reveal four findings. First, only matching marginal distribution is not sufficient for the cross-domain detection by comparing SSD+M with SSD+M+C (or SSD+M+WC); Second, only aligning conditional distribution is also not optimal because it lacks the unsupervised feature matching ensured by the marginal alignment, which can be validated by comparing SSD+C with SSD+M+C (or SSD+M+WC); Third, joint alignment can well alleviate the performance degradation caused by distribution discrepancies of samples; Fourth, safe transfer can be achieved by considering the transferability of each class, which can be shown by comparing SSD+M+C with SSD+M+WC.

Next, we compare the SSD+M+WC with state-of-the-art cross-domain detection methods [12], [16], [50]. DT-SSD [12] represents the pixel-level domain transfer method for UDA. WST-BSR-SSD [16] exploits weak self-training (WST) and background score regularization (BSR), and can reduce the domain gaps and improve the domain adaptability of the single-stage detector, such as SSD, effectively. ADDA-SSD [50] aligns the distribution of features by reducing the distribution discrepancy in the feature space. For fair comparison, we compare the SSD+M+WC with all SSD-based cross-domain methods, including DT-SSD, WST-BSR-SSD, ADDA-SSD. Note that the DT-PL-SSD [12] uses a combination method of unsupervised pixel-level adaptation (DT) and weakly-supervised adaptation (PL). In order to compare the unsupervised adaptation part of DT-PL-SSD with our method, we only report their results of DT-SSD.

Besides, we show the mAP values of two-stage detectors on the adaptation from PASCAL VOC to Clipart. Overall, the experimental results demonstrate that our method has superiority under UDA setting over other state-of-the-art methods including single-stage methods [12], [16], [50] and two-stage methods [14], [17], [20]. Further, for the challenging UFDA setting, which is a meaningful task but still not be studied so far, our JADF also achieves exciting results.

Furthermore, since our method focuses on learning the generalized backbone between domains, the adapted detector should be effective on both the source and target domains. Therefore, we inspect the performance of the adapted detector on the source domain. It can be observed from Table I that after the adaptation for the target domain is achieved, our method still maintains its original performance on the source domain.

**Inference time:** We show the inference speed of both single-stage and two-stage detectors in Table I. The inference speed is evaluated with batch size 1 on CUDA 8.0, NVIDIA Titan X. Compared with the family of two-stage detectors, the single-stage detectors achieve shorter inference times, operating at 46 frames per second (FPS).

*2) PASCAL VOC to Comic and Watercolor:* Tables II and III report the results of adaptation from PASCAL VOC to Comic and Watercolor datasets, respectively. It can be seen that the JADF strengthens the transferability of the baseline detector for the comic and watercolor scenarios.

TABLE II
ADAPTATION RESULTS FROM PASCAL VOC TO COMIC. THE DEFINITION OF M, C, WC FOLLOWS TABLE I.

| Method | bike | bird | car | cat | dog | person | mAP |
|---|---|---|---|---|---|---|---|
| SSD [33] | 21.7 | 12.8 | 34.4 | 11.0 | 14.6 | 44.4 | 23.1 |
| UDA Setting | | | | | | | |
| SSD+M (ours) | 40.2 | 16.2 | 25.2 | 14.4 | 23.4 | 49.2 | 28.1 |
| SSD+C (ours) | 40.3 | 11.5 | 30.5 | 14.2 | 26.0 | 45.1 | 28.0 |
| SSD+M +C (ours) | 45.3 | **17.6** | 19.7 | **25.2** | **28.0** | 46.4 | 30.4 |
| SSD+M+WC (ours) | 47.6 | 14.9 | **33.4** | 24.7 | 24.8 | **53.2** | **32.6** |
| DT-SSD [12] | 43.6 | 13.6 | 30.2 | 16.0 | 26.9 | 48.3 | 29.8 |
| WST-BSR-SSD [16] | **50.6** | 13.6 | 31.0 | 7.5 | 16.4 | 41.4 | 26.8 |
| ADDA-SSD [50] | 39.5 | 9.8 | 17.2 | 12.7 | 20.4 | 43.3 | 23.8 |
| UFDA Setting (n=3) | | | | | | | |
| SSD+M (ours) | 40.1±1.3 | 13.1±0.7 | 25.8±4.5 | 15.9±0.9 | 19.3±1.6 | 46.5±1.2 | 26.8±0.8 |
| SSD+C (ours) | 37.3±2.0 | 13.6±0.9 | 29.5±0.5 | 11.4±1.3 | 18.0±1.8 | 43.2±0.8 | 25.5±1.0 |
| SSD+M+C (ours) | 41.3±2.6 | **16.0**±0.9 | 26.9±2.6 | **18.5**±2.2 | **23.0**±0.8 | 48.2±1.7 | 29.0±0.7 |
| SSD+M+WC(ours) | **46.3**±1.1 | 15.1±0.7 | **31.5**±0.9 | 17.3±2.7 | 21.6±0.5 | **52.9**±0.9 | **30.8**±0.6 |

TABLE III
ADAPTATION RESULTS FROM PASCAL VOC TO WATERCOLOR. THE DEFINITION OF M, C, WC FOLLOWS TABLE I.

| Method | bike | bird | car | cat | dog | person | mAP |
|---|---|---|---|---|---|---|---|
| SSD [33] | 65.8 | 44.0 | 46.0 | 26.7 | 27.7 | 60.5 | 45.1 |
| UDA Setting | | | | | | | |
| SSD+M (ours) | 77.4 | 50.2 | 46.9 | 35.1 | 34.6 | 64.9 | 51.5 |
| SSD+C (ours) | 76.5 | 46.2 | 44.7 | 34.5 | 36.9 | 62.1 | 50.2 |
| SSD+M +C (ours) | **86.4** | 47.7 | 45.2 | 36.7 | 34.1 | 64.1 | 52.4 |
| SSD+M+WC (ours) | 83.8 | **51.2** | 47.6 | **38.2** | **38.5** | **65.7** | **54.2** |
| DT-SSD [12] | 82.8 | 47.0 | 40.2 | 34.6 | 35.3 | 62.5 | 50.4 |
| WST-BSR-SSD [16] | 75.6 | 45.8 | **49.3** | 34.1 | 30.3 | 64.1 | 49.9 |
| ADDA-SSD [50] | 79.9 | 49.5 | 39.5 | 35.3 | 29.4 | 65.1 | 49.8 |
| UFDA Setting (n=3) | | | | | | | |
| SSD+M (ours) | 78.9±2.9 | 48.5±1.1 | 41.2±1.2 | 34.1±1.3 | 31.7±1.6 | 64.3±1.3 | 49.8±1.7 |
| SSD+C (ours) | 73.7±2.5 | 48.1±0.7 | 42.1±2.0 | 32.0±0.8 | 31.8±2.6 | 63.5±1.2 | 48.5±1.0 |
| SSD+M+C (ours) | 75.9±1.6 | 48.2±0.5 | 43.8±0.6 | 33.3±1.0 | 33.4±1.2 | 65.2±0.6 | 50.0±0.6 |
| SSD+M+WC(ours) | **86.2**±1.0 | **48.9**±0.6 | **45.5**±0.3 | **34.5**±1.1 | **34.0**±1.9 | **67.2**±1.6 | **52.7**±0.6 |

TABLE IV
ADAPTATION RESULTS FROM CITYSCAPES TO FOGGYCITYSCAPES EMPLOYING ANOTHER SINGLE-STAGE DETECTOR (REFINEDET [37]) UNDER THE UDA SETTING. THE DEFINITION OF M, C, WC FOLLOWS TABLE I. REFINEDET ORACLE AND FASTER ORACLE REFER TO TRAINING THE DETECTOR ON THE LABELED TARGET IMAGES.

| | Method | bus | bicycle | car | bike | prsn | rider | train | truck | mAP on Target | FPS |
|---|---|---|---|---|---|---|---|---|---|---|---|
| Single-stage Detector | RefineDet [37] | 24.3 | 28.9 | 38.0 | 21.8 | 26.1 | 28.5 | 8.3 | 6.6 | 22.8 | |
| | RefineDet+M (ours) | 30.5 | 33.1 | 45.1 | 27.8 | 30.9 | 33.6 | 23.4 | 11.7 | 29.5 | |
| | RefineDet+M+C (ours) | 34.4 | 33.8 | 52.9 | 29.7 | 33.1 | 37.0 | 26.5 | 18.0 | 33.1 | |
| | RefineDet+M+WC (ours) | 33.5 | **34.1** | 53.7 | 30.1 | 33.4 | 38.5 | 34.3 | 18.3 | 34.5 | **24.1** |
| | DT-RefineDet [12] | 30.0 | 30.6 | 43.5 | 22.8 | 30.8 | 32.9 | 20.0 | 11.5 | 27.8 | |
| | ADDA-RefineDet [50] | 25.1 | 29.1 | 39.6 | 22.0 | 23.8 | 30.7 | 22.8 | 10.9 | 25.5 | |
| | RefineDet Oracle | 41.9 | 38.7 | 63.3 | 33.3 | 39.9 | 42.8 | 34.8 | 24.3 | 39.8 | |
| Two-stage Detector | Faster [29] | 22.3 | 26.5 | 34.3 | 15.3 | 24.1 | 33.1 | 3.0 | 4.1 | 20.3 | |
| | DA-Faster [14] | 25.0 | 31.0 | 40.5 | 22.1 | **35.3** | 20.2 | 20.0 | 27.1 | 27.6 | |
| | MAF [15] | 39.9 | 33.9 | 43.9 | 29.2 | 28.2 | 39.5 | 33.3 | 23.8 | 34.0 | 7.0 |
| | SW-Faster [17] | 36.2 | **35.3** | 43.5 | 30.0 | 29.9 | 42.3 | 32.6 | 24.5 | 34.3 | |
| | SW-Faster-ICR-CCR [20] | **45.1** | 34.6 | 49.2 | 30.3 | 32.9 | **43.8** | 36.4 | 27.2 | 37.4 | |
| | Faster Oracle | 51.9 | 37.8 | 53.0 | 36.8 | 36.2 | 47.7 | 41.0 | 34.7 | 42.4 | |

Next, the proposed SSD+M+WC is compared with SSD-based cross-domain methods including DT-SSD [12], WST-BSR-SSD [16], ADDA-SSD [50]. The experimental results demonstrate that the SSD+M+WC exceeds all the SSD-based cross-domain methods with a considerable margin, further validating the effectiveness of our JADF for cross-domain object detection.

*3) A Real-world Application:* In the following, we evaluate the domain adaptability of the proposed JADF under different weather conditions. We follow the official split lists in [14], [15], [17], which adopts *train* list (containing 2975 labeled images) of Cityscapes dataset as the source domain and *train* list (containing 2975 unlabeled images) of FoggyCityscapes dataset as the target domain. The experimental evaluation is conducted on the official *validation* set of FoggyCityscapes dataset.

For cross-domain adaptation from Cityscapes to FoggyCityscapes, we select RefineDet [37] as a new baseline model to further validate that JADF can be generalized to different detectors, where RefineDet+M+WC denotes the RefineDet equipped with all developed modules.

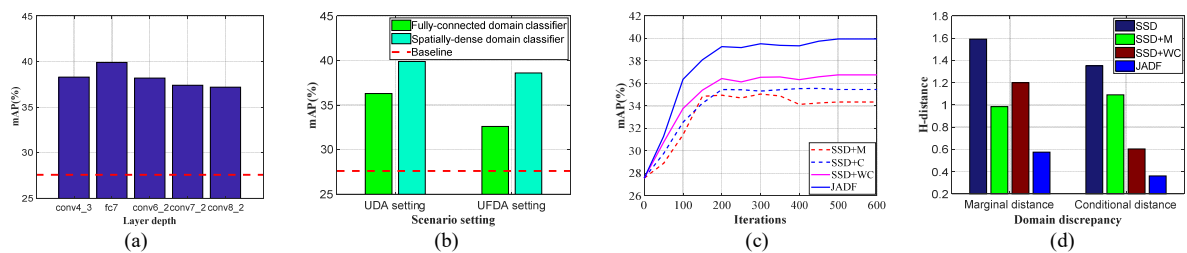

Fig. 4. Analyses of using the JADF to adapt SSD baseline to a new target domain, by aligning (a) different layers in SSD, and by employing (b) a spatially-dense domain classifier. Besides, (c) and (d) represent the convergence curve and $\mathcal{H}$-distance of different models, respectively.

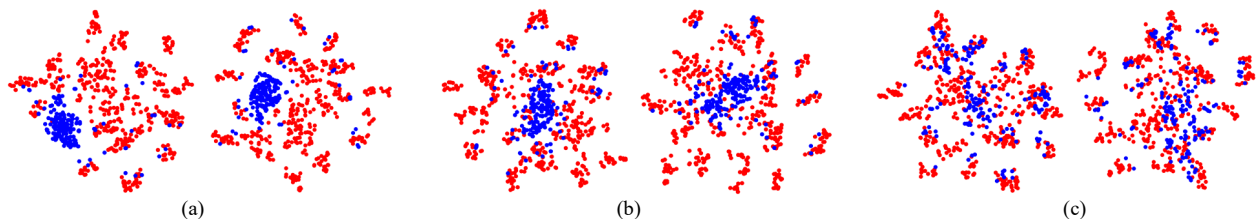

Fig. 5. Visualization of features with different adaptation modules, where (a), (b), and (c) represent the features extracted from SSD, SSD+M, SSD+M+WC, respectively, to align the fc7 layer (left) and the Conv6_2 layer (right). The red and blue points denote the features from the source and the target domains, respectively.

TABLE V
CHANGING THE NUMBER OF UNLABELED TARGET IMAGES.

| Method | One Shot | Two Shot | Three Shot | UDA Setting |
|---|---|---|---|---|
| SSD | 27.6 | 27.6 | 27.6 | 27.6 |
| SSD+M | 30.7±0.2 | 32.3±0.4 | 33.2±0.8 | 34.3 |
| SSD+C | 32.1±0.8 | 33.6±0.6 | 34.6±0.7 | 35.5 |
| SSD+WC | 33.8±0.5 | 35.0±0.8 | 36.2±0.7 | 36.7 |
| SSD+M+WC | 35.9±1.2 | 36.9±0.7 | 38.5±0.2 | 39.9 |

Table IV demonstrates the comparison results. It can be observed that the proposed RefineDet+M+WC outperforms all the single-stage detectors with a considerable margin in terms of mAP value. Another observation from Table IV is that the mAP of the proposed RefineDet+M+WC is lower than that of SW-Faster-ICR-CCR [20]. This is mainly because SW-Faster-ICR-CCR [20] uses a domain adaptive detector (SW-Faster [17]) as the baseline detector that can achieve mAP of 34.8% on FoggyCityscapes. Actually, the ICR-CCR module proposed in [20] contributes only 2.6% for the performance gain. By contrast, the proposed RefineDet+M+WC boosts the performance of baseline from 22.8% to 34.5%, indicating that the proposed JADF achieves a considerable performance improvement.

*E. Insight Analyses*

*1) Ablation Study:* We conduct the ablation studies from three aspects. First, the capability of cross-domain detection when changing the number of unlabeled target images $n$ under the UFDA setting; Second, the impact of adapting different layers of the baseline detector on mAP; Third, the impact of employing the spatially-dense domain classifier on mAP. For the first item, Table V shows the relationship between the number of unlabeled target samples and the performance of cross-domain detection. For the second item, we report the results of aligning different prediction layers for SSD [33] (from Conv4_3 to Conv8_2) and find that the performance of aligning fc7 layer is optimal, as shown in Fig. 4(a). This is mainly because high-level features in SSD lose more spatial information. For the last item, as analyzed in Section III, since there may be multiple instances in an input image, we employ a spatially-dense domain classifier to perform the patch-level feature adaptation that can fully consider multiple potential objects. Fig. 4(b) indicates performance improvement when using the spatially-dense classifier. This implies that for domain adaptation detection, a patch-level feature adaptation is helpful, especially in the UFDA setting.

*2) Convergence Speed and Distribution Discrepancy:* The proposed JADF has faster convergence speed than other baseline detectors, as shown in Fig. 4(c). As described in Section III, the $\mathcal{H}$-distance can measure the discrepancy for different distributions. Fig. 4(d) shows the $\mathcal{H}$-distance of marginal and conditional distributions of features, and indicates that JADF can reduce the discrepancies of the marginal and conditional distributions of features more effectively.

*3) Feature and Example Visualization:* We visualize the features of adapting from PASCAL VOC to Clipart by SSD, SSD-M and SSD+M+WC (aligning the fc7 layer in our implementation) in Fig. 5 by means of t-SNE [53]. The visualizations show that the features can be better aligned by using JADF. Besides, more visual examples of detection results on the target domain are shown in Figs. 6 and 7.

V. CONCLUSION

In this paper, we have proposed JADF, a simple and easy-to-utilize, end-to-end adaptive detection framework,

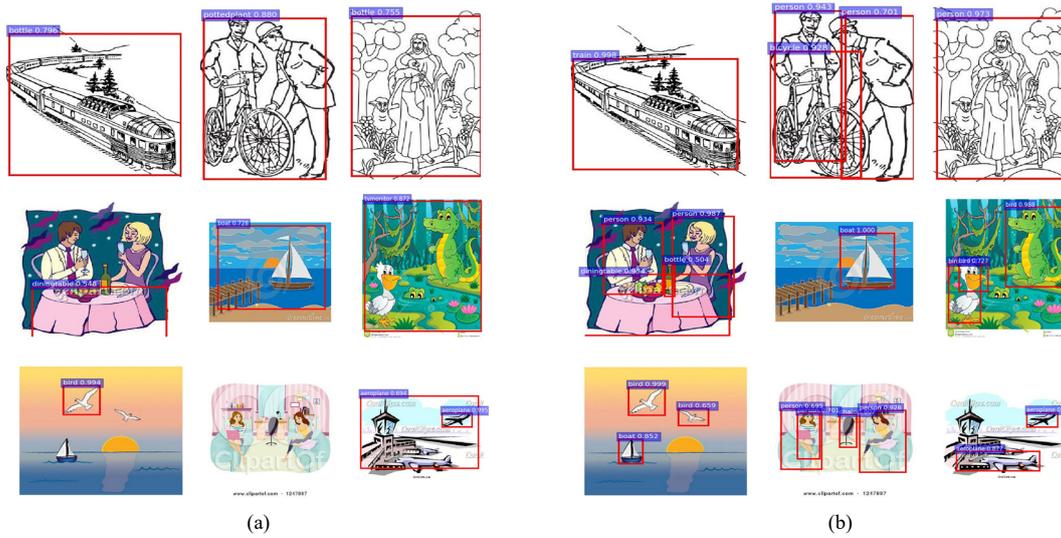

Fig. 6. Visual detection results for adapting SSD and JADF from PASCAL VOC to Clipart. (a) Detection results predicted by the SSD detector. (b) Detection results predicted by JADF.

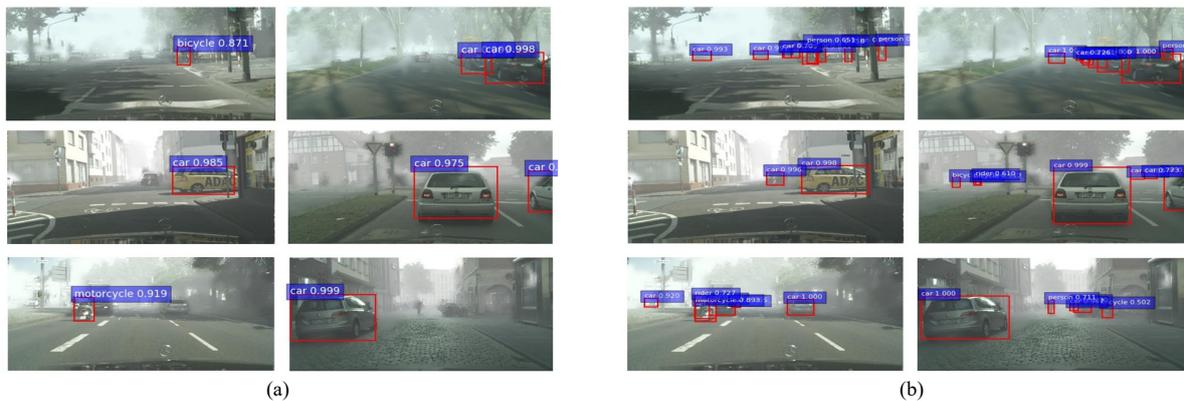

Fig. 7. Visual detection results for adapting SSD and JADF from Cityscapes to FoggyCityscapes. (a) Detection results predicted by the SSD detector. (b) Detection results predicted by JADF.

towards transferring label-rich knowledge on the source domain to an unlabeled target domain. JADF considers both marginal and conditional distribution alignment between domains and performs the joint adaptation process through adversarial learning. Besides, since different object classes often have different domain semantic gaps during the conditional adaptation process, we further propose a class-wise transferability assessment to weaken the cross-domain alignment of hard-to-transfer classes and strengthen the cross-domain matching of easy-to-transfer classes. The experimental results have demonstrated that JADF significantly exceeds the state-of-the-art cross-domain detection methods in UDA setting and achieves exciting results in the challenging UFDA setting.

Since the object annotations for the target domain are unavailable, we have to use a well-trained source detector to approximately predict them in the conditional adaptation process. This approximation may result in prediction deviation for the object annotations. Thus, how to provide an unbiased prediction for the target domain needs to be further studied in the future.